\documentclass[sigconf]{acmart}
\AtBeginDocument{%
  \providecommand\BibTeX{{%
    \normalfont B\kern-0.5em{\scshape i\kern-0.25em b}\kern-0.8em\TeX}}}

\copyrightyear{2021}
\acmYear{2021}
\setcopyright{acmcopyright}\acmConference[MM '21]{Proceedings of the 29th ACM
International Conference on Multimedia}{October 20--24, 2021}{Virtual Event, China}
\acmBooktitle{Proceedings of the 29th ACM International Conference on Multimedia
(MM '21), October 20--24, 2021, Virtual Event, China}
\acmPrice{15.00}
\acmDOI{10.1145/3474085.3475592}
\acmISBN{978-1-4503-8651-7/21/10}

\begin{document}
\fancyhead{}
\title{Visible Watermark Removal via Self-calibrated Localization and Background Refinement}

\author{Jing Liang}
\affiliation{%
 \institution{MoE Key Lab of Artificial Intelligence, \\
  Shanghai Jiao Tong University}
}
\email{leungjing@sjtu.edu.cn}

\author{Li Niu}
\authornote{Corresponding Author}
\affiliation{
 \institution{MoE Key Lab of Artificial Intelligence,\\
  Shanghai Jiao Tong University
  }
}
\email{ustcnewly@sjtu.edu.cn}
\author{Fengjun Guo}
\affiliation{
  \institution{INTSIG}
}
\email{fengjun_guo@intsig.net}

\author{Teng Long}

\affiliation{
  \institution{INTSIG
}
}
\email{mike_long@intsig.net}

\author{Liqing Zhang}
\affiliation{
  \institution{MoE Key Lab of Artificial Intelligence,\\
  Shanghai Jiao Tong University
  }
}
\email{zhang-lq@cs.sjtu.edu.cn}


\begin{abstract}
  Superimposing visible watermarks on images provides a powerful weapon to cope with the copyright issue. Watermark removal techniques, which can strengthen the robustness of visible watermarks in an adversarial way, have attracted increasing research interest. Modern watermark removal methods perform watermark localization and background restoration  simultaneously, which could be viewed as a multi-task learning problem. However, existing approaches suffer from incomplete detected watermark and degraded texture quality of restored background. Therefore,  we design a two-stage multi-task  network to address the above issues. The coarse stage consists of a watermark branch and a background branch, in which the watermark branch self-calibrates the roughly estimated mask and passes the calibrated mask to background branch to reconstruct the watermarked area. In the refinement stage, we integrate multi-level features to improve the texture quality of watermarked area. Extensive experiments on two datasets demonstrate the effectiveness of our proposed method. 
\end{abstract}

\keywords{watermark removal; multi-task learning; two-stage network}

\ccsdesc[500]{Computing methodologies~Image processing}

\maketitle

\section{Introduction}
With the surge of social media, images become the most prevailing carriers for recording and conveying information. To protect the copyright or claim the ownership, various types of visible watermarks are designed and overlaid on background images via alpha blending. Superimposing visible watermark is considered as an efficient and effective approach to combat against attackers. However, watermarked images are likely to be converted back to watermark-free images by virtue of modern watermark removal techniques. To evaluate and strengthen the robustness of visible watermarks in an adversarial way, watermark removal task has raised the research interest in recent years~\cite{dekel2017effectiveness, cheng2018large, li2019towards, cao2019generative, hertz2019blind,cun2020split}.

\begin{figure}[t]
\centering
\includegraphics[width=0.42\textwidth]{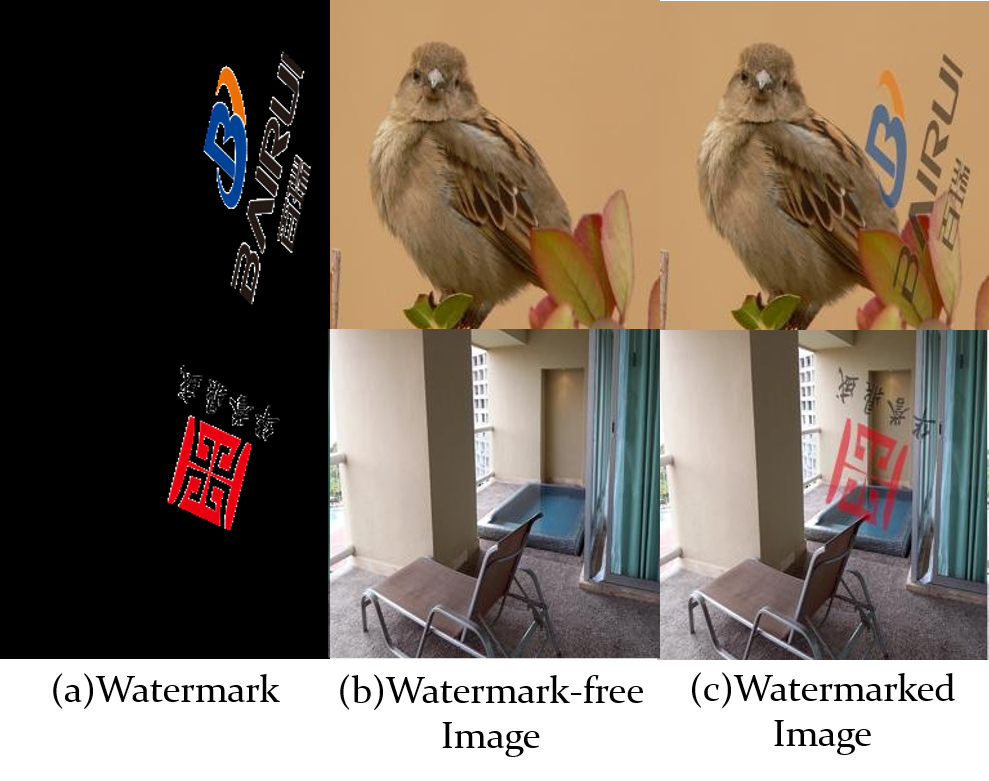}
\caption{The watermarked image (c) is acquired by superimposing a watermark (a) on the background image (b) via alpha blending. Given a watermarked image (c),  watermark removal task aims to reconstruct the watermark-free image (b) without knowing the watermark mask. }
\label{fig:teaser}
\end{figure}

Watermark removal, which aims to reconstruct  background images based on watermarked images, is an open and challenging problem. Watermarks can be overlaid at any position of a background image with different sizes, shapes, colors, and transparencies. Besides, the watermarks often contain complex patterns like warped symbols, thin line, shadow effects, \emph{etc.} The above reasons render the watermark removal task dramatically difficult when no prior knowledge is provided. An example of watermark, watermark-free image, and watermarked image  is shown in Figure ~\ref{fig:teaser}. In the remainder of this paper, we use two terms ``background image" and ``watermark-free image" exchangably.

In some pioneer works, the location of watermarked area is required. Guided by watermark mask, watermark removal is similar to image inpainting~\cite{huang2004attacking} or feature matching problem~\cite{park2012identigram, pei2006novel}. Nevertheless, manually annotating the watermark mask for each image is extremely time-consuming and cost-expensive. Noticing the fact that multiple images are often marked with the same watermark,  watermark could be detected and removed in a more effective way~\cite{dekel2017effectiveness, gandelsman2019double}. Unfortunately, the assumption in \cite{dekel2017effectiveness, gandelsman2019double} limits their application to real-world scenarios. 
Recently, researchers~\cite{cheng2018large, li2019towards, cao2019generative, hertz2019blind,cun2020split, liu2021wdnet} attempt to solve the blind watermark removal problem in an end-to-end manner with deep learning approaches. Some works~\cite{li2019towards,cao2019generative} formulated the watermark removal problem as an image-to-image translation task without localizing the watermark.  On the contrary, other works realized that watermark should be localized and removed sequentially~\cite{cheng2018large} or simultaneously~\cite{hertz2019blind,cun2020split, liu2021wdnet}.  Despite the great success of these emerging methods, they are still struggling to localize the watermark precisely and completely, especially when the watermark has complex patterns, diverse colors, or isolated fragments.
The inaccurate watermark mask will interfere the reconstruction of background image. Moreover, the reconstructed images are suffering from quality issues like blur, artifacts, and distorted structures, which awaits further improvement.

In this paper, we propose a novel watermark removal network via \textbf{S}elf-calibrated \textbf{L}ocalization and \textbf{B}ackground \textbf{R}efinement (\textbf{SLBR}), which consists of a coarse stage and a refinement stage. In the coarse stage, we consider watermark localization and watermark removal as two tasks in a multi-task learning framework. Specifically, we employ a U-Net ~\cite{ronneberger2015u}  structure, in which two tasks share the same encoder but have two separate decoders. The mask decoder branch predicts multi-scale watermark masks, which provides guidance for the background decoder branch via Mask-guided Background Enhancement (MBE) module to better reconstruct watermark-free images. Considering that the watermarks in various images are considerably different in many aspects, we design a Self-calibrated Mask Refinement (SMR) module, in which the watermark feature is propagated to the whole feature map to better handle image-specific watermark. 
In the refinement stage, we take the predicted watermark mask and watermark-free image in the coarse stage as input, to produce a refined watermark-free image. To fully exploit the useful information in the coarse stage, we add skip-stage connections between the background decoder branch in the coarse stage and the encoder in the refinement stage. Considering that different levels of features capture the structure information or texture details, we repeatedly use Cross-level Feature Fusion (CFF) modules to aggregate multi-level encoder features in the refinement stage. The output image from the refinement stage is the final recovered background image. Our main contributions could be summarized as follows,
\begin{itemize}
    \item We propose a novel two-stage multi-task network named SLBR with cross-stage and cross-task information propagation for watermark removal task.
    \item In the coarse stage, we devise a novel Self-calibrated Mask Refinement (SMR) module to calibrate the watermark mask and a novel Mask-guided Background Enhancement (MBE) module to enhance the background representation. 
    \item In  the refinement stage, we propose a novel  Cross-level Feature Fusion (CFF) module, which is repeatedly used to get the refined watermark-free image.
    \item Experiments on two datasets demonstrate the effectiveness of our proposed method.
\end{itemize}

\section{Related Works}
In this section, we first introduce a broad range of image content removal applications, and then describe the existing watermark removal methods. Besides, since our network involves multi-level feature fusion, we also briefly review the related methods.

\noindent\textbf{Image Content Removal: }Similar to watermark removal task, some existing tasks also focus on removing some undesirable content from an image, for example, deraining ~\cite{fan2018residual,qian2018attentive, yang2019single, wang2020joint}, blind shadow removal ~\cite{wang2018stacked, ding2019argan, cun2020towards}, dehazing ~\cite{yang2018proximal, zhang2018densely, cai2016dehazenet, he2010single, dong2020multi, cong2020discrete, zhang2020nighttime}, and so on. However, these removed contents (\emph{e.g.}, rain, shadow, haze) often consist of repeated patterns and monotonous colors. Different from the above tasks, watermark removal task targets at removing the watermarks which have diverse shapes and colors. Therefore, watermark removal task is a unique and challenging task.

\noindent\textbf{Visible Watermark Removal: } Visible watermark provides a powerful weapon for protecting the copyright. To evaluate and improve the robustness of visible watermarks, watermark removal techniques are proposed and gradually draw attentions from the security community. In the earlier explorations~\cite{huang2004attacking, park2012identigram, pei2006novel}, they generally interacted with users to indicate the watermark locations for the following background recovery, which limits its practical usage. Since acquiring each of image location is ineffective,  ~\cite{dekel2017effectiveness, gandelsman2019double} assumed that  multiple images have the same watermark pattern, in which multiple images are processed simultaneously to remove the common watermark pattern. However, the assumption in~\cite{dekel2017effectiveness, gandelsman2019double}  is too stringent and unpractical, which weakens its potential in real-world applications.

The development of deep learning techniques have greatly advanced the watermark removal task. Some methods~\cite{li2019towards, cao2019generative} formulated the watermark removal as an image-to-image translation task. Other methods~\cite{hertz2019blind, cun2020split, liu2021wdnet} performed watermark localization and removal tasks at the same time.   In~\cite{hertz2019blind,cun2020split, liu2021wdnet}, watermark localization and watermark removal were wrapped up in a multi-task learning framework. Nevertheless, the above methods~\cite{li2019towards, cao2019generative, hertz2019blind,cun2020split, liu2021wdnet} are still struggling to achieve satisfactory performance in localizing watermark and restoring the watermark-free images.

\noindent\textbf{Multi-level Feature Fusion: }Multi-level feature fusion has been widely used in various computer vision tasks~\cite{jiang2020multi, dong2020multi, cun2020towards, zhang2017amulet, liu2018path} for boosting network performance. Aggregation strategies could vary from task to task, but most of them fall into the following classical scopes: dense connection~\cite{zhang2018multi}, top-down and/or bottom up feature integration~\cite{liu2018path, lin2017feature}, feature concatenation~\cite{zhang2018residual, cun2020towards, zhang2017amulet}, weighted element-wise summation~\cite{chen2019gated, zhang2018bi}. Although these methods are capable of merging multi-level features, how to propagate  multi-level information properly and efficiently in watermark removal task is still unsolved.  
In watermark removal approaches~\cite{hertz2019blind, liu2021wdnet}, Hertz et al.~\cite{hertz2019blind} only considered the skip connection from encoder; Liu et al.~\cite{liu2021wdnet} further passed the shallowest decoder feature from coarse stage to refinement stage. Nevertheless, these methods overlook the potential capacity of multi-level features integration. Thus, we propose to bridge the coarse stage and refinement stage by multi-level feature propagation, and further perform cross-level feature interweaving for better background reconstruction.

\section{Our Method}

\begin{figure*}[t]
    \centering
    \includegraphics[width=0.95\textwidth]{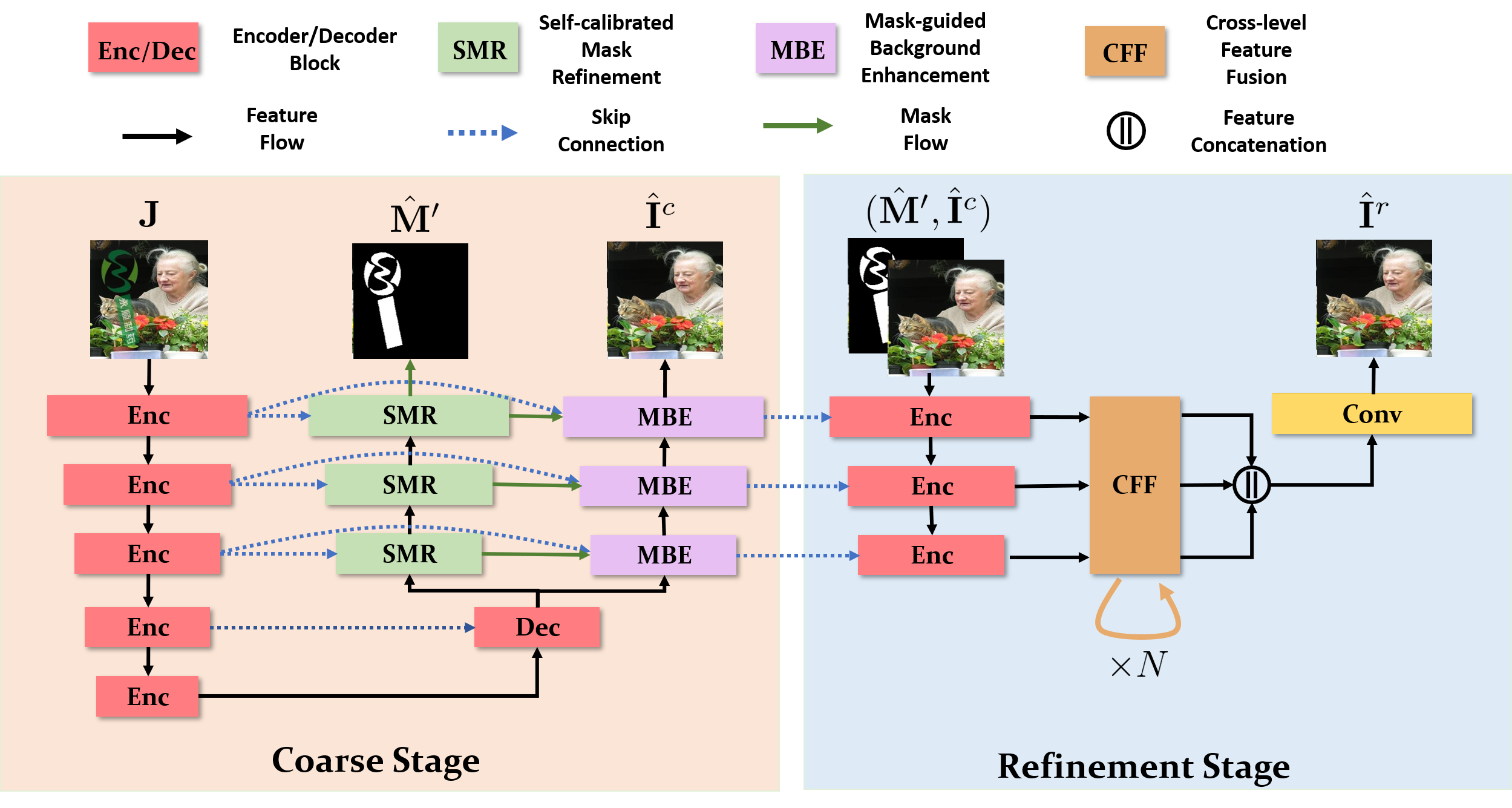}
    \caption{ The illustration of our SLBR network  which consists of a coarse stage and a refinement stage. The coarse stage contains one shared encoder and two separate decoder branches, which accounts for watermark localization and watermark-free image reconstruction respectively. The refinement stage takes the predicted watermark mask and watermark-free image from the coarse stage, producing the refined watermark-free image. We omit the side output masks in this figure for clarity.}
    \label{fig:framework}
\end{figure*}

Given a watermarked image $\mathbf{J}$ which is obtained by superimposing a watermark on the background image $\mathbf{I}$, the goal of watermark removal is recovering the watermark-free image $\mathbf{I}$ based on the watermarked image $\mathbf{J}$. Because the watermark mask $\mathbf{M}$ is usually unknown, we need to perform two tasks simultaneously: watermark localization and watermark removal, which can be accommodated under a multi-task learning framework. As exemplified in Figure~\ref{fig:framework}, our whole network is designed in a coarse-to-fine manner, which comprises of a coarse stage and a refinement stage. In the coarse stage, similar to previous multi-task learning methods~\cite{liu2021wdnet, hertz2019blind}, we employ one shared encoder and two split decoders, in which two decoders account for localizing the watermark (mask decoder branch) and restoring the background image (background decoder branch) respectively. In the mask decoder branch, we design a Self-calibrated Mask Refinement (SMR) module to promote the quality of predicted watermark mask. To ease the information flow from the mask decoder branch to the background decoder branch, we employ a Mask-guided Background Enhancement (MBE) module to enhance the background decoder features. In the refinement stage, we build skip-stage connections between the decoder features in the coarse stage and the encoder features in the refinement stage to facilitate information propagation from coarse stage to refinement stage. To better recover the structure and texture  of background image, we also devise a Cross-level Feature Fusion (CFF) module to aggregate multi-level encoder features iteratively in the refinement stage. Next, we will elaborate on the coarse stage in Section~\ref{sec:coarse_stage} and the refinement stage in Section~\ref{sec:fine_stage}.

\subsection{Coarse Stage} \label{sec:coarse_stage}
In the coarse stage, we adopt the U-Net~\cite{ronneberger2015u} architecture with skip links connecting encoder and decoder features as shown in Figure~\ref{fig:framework}.  
Specifically, we employ the structure of encoder block and decoder block in \cite{hertz2019blind}.
Watermark localization and watermark removal are treated as two tasks, which share all five encoder blocks and the first decoder block. But they have three separate decoder blocks, which form the mask decoder branch and background decoder branch separately.
In the mask decoder branch,  it is equipped with our designed Self-calibrated Mask Refinement (SMR) module and assigned to indicate watermark position. Apart from the predicted mask from the last decoder block, we also predict side output masks based on the features in the other two decoder blocks. In the background decoder branch, it is composed of Mask-guided Background Enhancement (MBE) module and assigned to recover the corrupt background area overlaid with watermark. SMR and MBE block will be detailed next.


\begin{figure*}[!ht]
    \centering
    \includegraphics[width=1.0\textwidth]{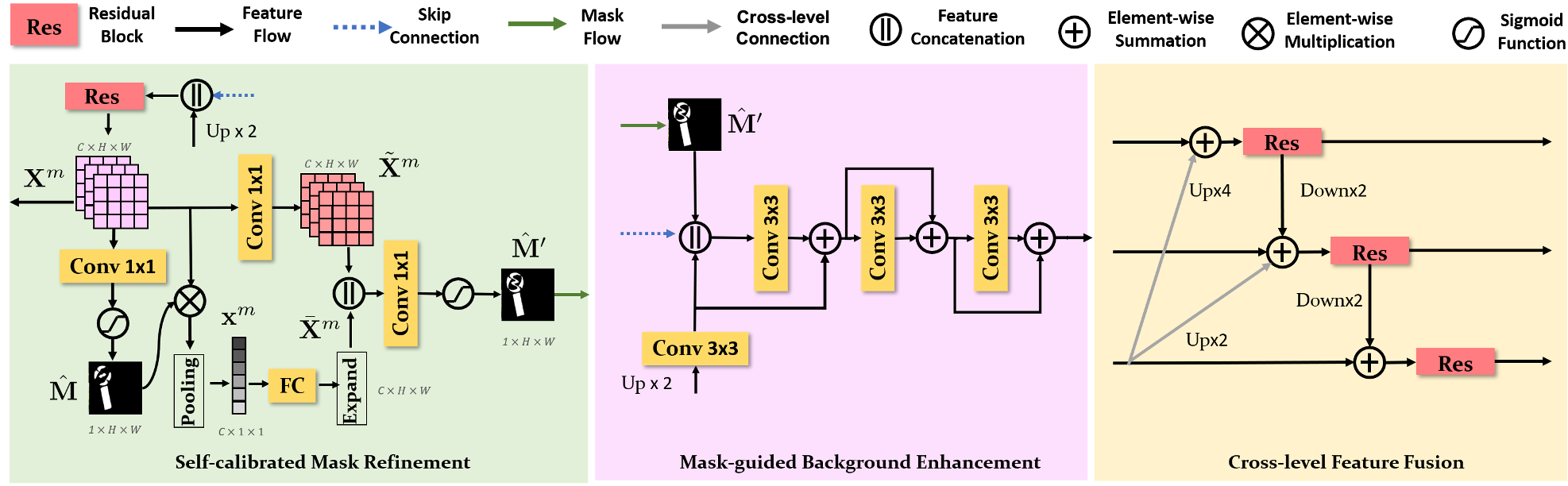}
    \caption{ The illustration of our Self-calibrated Mask Refinement (SMR), Mask-guided Background Enhancement (MBE), Cross-level Feature Fusion (CFF) modules. ``Pooling" means average pooling. ``FC" means fully-connected layer. ``Expand" means spatial replication. }
    \label{fig:modules}
\end{figure*}

\noindent\textbf{Self-calibrated Mask Refinement (SMR) module: }When predicting the watermark mask, we observe that the predicted masks are often incomplete. One possible reason is that the watermarks in different images have diverse shapes, colors, patterns, and transparencies, so one global predictor can hardly localize all various types of watermarks. Thus, we consider calibrating the mask predictor according to the watermark characteristics in each image, to improve the quality of predicted watermark mask. By taking the last decoder block in the mask decoder branch as an example, as shown in Figure~\ref{fig:modules}, we concatenate the features from previous decoder block and skip connection, followed by stacked residual blocks~\cite{liu2021wdnet}. We denote that $\mathbf{X}^m$ is the feature map used to predict the watermark mask $\hat{\mathbf{M}}$. Following~\cite{liu2021wdnet, hertz2019blind}, we use binary cross-entropy loss to enforce $\hat{\mathbf{M}}$ to be close to the ground-truth watermark mask $\mathbf{M}$: 
\begin{eqnarray}\label{eqn:L_mask}
\mathcal{L}_{mask}= - \sum_{i,j} \left( M_{i,j} \log \hat{M}_{i,j} + (1-M_{i,j}) \log (1-\hat{M}_{i,j}) \right),
\end{eqnarray}
where $M_{i,j}$ (\emph{resp.}, $\hat{M}_{i,j}$) is the $(i,j)$-th entry in $\mathbf{M}$(\emph{resp.}, $\hat{\mathbf{M}}$).
We first apply this roughly estimated mask $\hat{\mathbf{M}}$ to the feature map ${\mathbf{X}^m}$ to pool the averaged feature vector ${\mathbf{x}^{m}}$. Although the estimated mask $\hat{\mathbf{M}}$ has missed detection and false alarms, watermarked pixels still dominate the estimated mask and thus the averaged feature vector can roughly represent the watermark characteristics. After obtaining the averaged watermark feature ${\mathbf{x}^{m}}$, we tend to compare all pixel-level features in $\mathbf{X}^m$ with ${\mathbf{x}^{m}}$. 
Specifically, we first employ a $1\times 1$ conv layer (\emph{resp.}, fully-connected layer) to project $\mathbf{X}^m$ (\emph{resp.}, ${\mathbf{x}^{m}}$) to $\tilde{\mathbf{X}}^m$ (\emph{resp.}, $\tilde{\mathbf{x}}^m$). In the projected space, we expect that the averaged watermark feature is close to the watermarked pixels but far away from the unmasked pixels. Then, we spatially replicate $\tilde{\mathbf{x}}^{m}$ to the same size as $\tilde{\mathbf{X}}^m$, giving rise to $\bar{\mathbf{X}}^m$. We concatenate $\tilde{\mathbf{X}}^{m}$ and $\bar{\mathbf{X}}^m$, followed by a $1\times 1$ conv layer to predict a binary affinity map, in which $1$ (\emph{resp.}, $0$) indicates that this pixel-level feature is similar (\emph{resp.}, dissimilar) to the averaged watermark feature. Apparently, the ground-truth affinity map should be identical with the ground-truth watermark mask. Therefore, we can apply the same loss as Eqn. (\ref{eqn:L_mask}) to supervise the affinity map. By using $\hat{\mathbf{M}}'$ to denote the predicted affinity map, the loss can be expressed as 
\begin{eqnarray}\label{eqn:L_mask_refine}
\mathcal{L}'_{mask}= - \sum_{i,j} \left( M_{i,j} \log \hat{M}'_{i,j} + (1-M_{i,j}) \log (1-\hat{M}'_{i,j}) \right),
\end{eqnarray}
in which $\hat{M}'_{i,j}$ is the $(i,j)$-th entry in $\hat{\mathbf{M}}'$.
By comparing all pixel-level features with the averaged watermark feature, the predicted affinity map can identify some missed detection and erase some false alarms. Because $\hat{\mathbf{M}}'$ is refined $\hat{\mathbf{M}}$, we use $\hat{\mathbf{M}}'$ as input for the background decoder branch and the refinement stage. We refer to the above module as Self-calibrated Mask Refinement (SMR) module and replace the original decoder blocks~\cite{hertz2019blind} in the mask decoder branch by our SMR modules.


\noindent\textbf{Mask-guided Background Enhancement (MBE) module: }In the coarse stage, watermark localization and watermark removal are two closely related tasks under a multi-task learning framework. According to~\cite{liu2021wdnet, cun2020split, hertz2019blind}, knowing the watermark area will offer strong guidance for the watermark removal task. In previous multi-task learning works ~\cite{dai2016instance,gao2019nddr,zhao2018modulation,liu2019end}, myriads of strategies have been proposed to encourage the information sharing and propagation across different tasks. In our problem, we conjecture that mask localization would provide more benefit for watermark removal than the other way around. Furthermore, our main goal is recovering the watermark-free image. Therefore, we design a Mask-guided Background Enhancement (MBE) module to guide the information flow from mask decoder branch to background decoder branch. 

As shown in Figure~\ref{fig:modules}, in each MBE module, we concatenate the output mask $\hat{\mathbf{M}}'$ from the corresponding SMR module with the features from previous decoder block and skip connection. Then, we apply a $3\times 3$ conv layer to the concatenated feature to generate a feature residue, which is added back to the input feature. Following~\cite{hertz2019blind}, We repeat this residual process for three times to produce the enhanced background decoder feature, which is fed into the next decoder block. 

We notice that in some previous multi-task learning networks~\cite{tsai2017deep,gu2020hard}, the features in one decoder branch are also appended to the features in the other decoder branch. Different from them, our MBE module incorporates the predicted mask and learns residual information to boost the capacity of background representation.  
Here, we denote generated background image as $\hat{\mathbf{I}}^c$, which is expected to be close to the ground-truth watermark-free image $\mathbf{I}$ using $L_1$ loss:
\begin{eqnarray} \label{eqn:L_bg_L1_coarse}
\mathcal{L}_{bg-{L_1}}^c= \|\mathbf{I} - \hat{\mathbf{I}}^c \|_1.
\end{eqnarray}

\subsection{Refinement Stage} \label{sec:fine_stage}
We observe that the restored watermark-free image $\hat{\mathbf{I}}^c$ in the coarse stage may suffer from some quality issues like blur, artifact, and distorted structure, which calls for further improvement. Thus, we additionally attach a refinement stage to the coarse stage. We concatenate the coarse watermark-free image $\hat{\mathbf{I}}^c$ and predicted watermark mask $\hat{\mathbf{M}}'$ as the input for the refinement stage. First, we employ three encoder blocks~\cite{hertz2019blind} to extract multi-level features. To fully exploit the repaired content information in the coarse stage, we add skip-stage connections between the decoder features in the coarse stage and the encoder features in the refinement stage. Although WDNet~\cite{liu2021wdnet} also uses the coarse-stage feature in the refinement stage, they simply append the last feature map in the coarse stage to the input for the refinement stage. Distinctively, we connect each background decoder feature in the coarse stage to its corresponding encoder feature with the same spatial size in the refinement stage in a symmetrical way, yielding enhanced multi-level encoder features in the refinement stage. Compared with~\cite{liu2021wdnet}, our skip-stage connections can integrate the content information in the coarse stage and refinement stage more thoroughly.

\noindent\textbf{Cross-level Feature Fusion (CFF) module: } Generally, we assume that the low-level encoder features with larger spatial size encode the texture details, while the high-level encoder features with smaller spatial size encode the structure information. To recover clear and coherent texture and structure for watermark-free image, we need to leverage multi-level encoder features in a better way. Thus, we design a Cross-level Feature Fusion (CFF) module, which is repeatedly used after the initial multi-level encoder features. As shown in Figure~\ref{fig:modules}, in each CFF module, we upsample the high-level encoder feature to the same size of different low-level encoder features. After concatenating the upsampled high-level encoder feature with each low-level encoder feature, we also apply stacked residual blocks~\cite{liu2021wdnet} to all encoder features including the high-level encoder feature. Besides this sparse connection fashion (\emph{i.e.},  only propagating the high-level feature to the other levels of features), we have also tried dense connection fashion (\emph{i.e.}, propagating all levels of features to the other levels of features) as in~\cite{zhang2017amulet}. However, we observe that sparse connection is able to achieve comparable or even better results than dense connection.
Thus, we adopt sparse connection in our CFF module for efficiency.
We stack CFF module for $N$ times ($N=3$ in our experiments). 

Finally, based on the multi-level encoder features output from the last CFF module, we resize the encoder features of all levels to the target image size and aggregate them to obtain the final feature map. A  $1\times 1$ conv layer is applied to the final feature map to generate the refined watermark-free image $\hat{\mathbf{I}}^r$. Similar to Eqn. (\ref{eqn:L_bg_L1_coarse}), we employ $L_1$ loss to enforce the refined watermark-free image to approach the ground-truth one:
\begin{eqnarray}
\mathcal{L}_{bg-{L_1}}^r =  \|\mathbf{I} - \hat{\mathbf{I}}^r\|_1.
\end{eqnarray}

To further ensure the quality of generated watermark-free image, we additionally employ perception loss~\cite{johnson2016perceptual, zhang2018unreasonable} based on VGG16 \cite{simonyan2014very} pretrained on ImageNet~\cite{deng2009imagenet}. The perception loss can be written as
\begin{eqnarray}
\mathcal{L}_{bg-vgg} =  \sum_{k\in 1,2,3} \|\Phi_{vgg}^{k}(\hat{\mathbf{I}}^r) - \Phi_{vgg}^{k}({\mathbf{I}})\|_1,
\end{eqnarray}
in which $\Phi_{vgg}^{k}(\cdot)$ means the activation map of $k$-th layer in VGG16. 

Finally, we collect the losses in the coarse stage and the refinement stage, leading to the total loss: 
\begin{eqnarray} \label{eqn:L_total}
\mathcal{L}_{all}= \!\!\!\!\!\!\!\!&&\mathcal{L}^c_{bg-{L_1}} + \mathcal{L}^r_{bg-{L_1}} + \lambda_{\text{vgg}} \mathcal{L}_{bg-vgg} + \nonumber\\
&&\lambda_\text{mask} (\mathcal{L}_\text{mask}+\mathcal{L}'_\text{mask}),
\end{eqnarray}
in which $\lambda_{\text{vgg}}$ and  $\lambda_\text{mask}$ are trade-off parameters. The whole network including the coarse stage and refinement stage can be trained in an end-to-end manner. In the testing stage, given a watermarked input image $\mathbf{J}$, we use the output image $\hat{\mathbf{I}}^r$ from the refinement stage as the final result.

\setlength{\tabcolsep}{12pt}
\begin{table*}[t] 
\begin{tabular}{c|c|c|c|c|c|c|c|c}
\toprule[1pt]
Method & \multicolumn{4}{c|}{LVW} & \multicolumn{4}{c}{CLWD} \\ \cline{2-9} 
 & PSNR $\uparrow$ & SSIM $\uparrow$  & RMSE $\downarrow$& RMSEw $\downarrow$ &  PSNR $\uparrow$ & SSIM  $\uparrow$& RMSE  $\downarrow$& RMSEw $\downarrow$ \\ \hline
U-Net~\cite{ronneberger2015u}       & 30.33         & 0.9517        & 7.11        & 42.18           &   23.21       & 0.8567     & 19.35         & 48.43    \\ 
Qian \emph{et al.}~\cite{qian2018attentive}           & 39.92        & 0.9902        & 3.31        & 21.40           & 34.60       & 0.9694      & 5.40          & 19.34   \\
Cun \emph{et al.}~\cite{cun2020towards}           & 40.68         & 0.9949        & 2.62        & 17.29           & 35.29         & 0.9712     & 5.28         & 18.25    \\ 
\hline
Li \emph{et al.}~\cite{li2019towards}           & 33.57         & 0.9690     & 5.84       & 34.71            & 27.96        & 0.9161    & 12.63          & 46.80     \\
Cao \emph{et al.}~\cite{cao2019generative}       & 34.16         & 0.9714        & 5.51        & 33.42           & 29.04         & 0.9363     & 10.36         & 41.21    \\ 
WDNet~\cite{liu2021wdnet}          & 42.45         & 0.9954        &  2.39        & 12.75          & 35.53      & 0.9738          & 5.11 &  17.27    \\
BVMR~\cite{hertz2019blind}         &  40.14         & 0.9910        & 3.24        & 18.57           & 35.89        & 0.9734     & 5.02          & 18.71     \\ 
SplitNet~\cite{cun2020split}         &  43.16        & 0.9946       & 2.28       & 14.06           & 37.41        & 0.9787    & 4.23          & 15.25     \\ \hline
\textbf{SLBR (Ours)}          & \textbf{43.48}         & \textbf{0.9959}       & \textbf{2.15}        & \textbf{12.14}           & \textbf{38.28}         & \textbf{0.9814}   & \textbf{3.76}          & \textbf{14.07}       \\ \bottomrule[1pt]
\end{tabular}
\caption{The results of different methods on LVW~\cite{cheng2018large} and CLWD~\cite{liu2021wdnet}  datasets. The best results are denoted in boldface.}
\label{tab:exp_baseline}
\end{table*}

\begin{figure*}[t]
    \centering
    \includegraphics[width=1.0\textwidth]{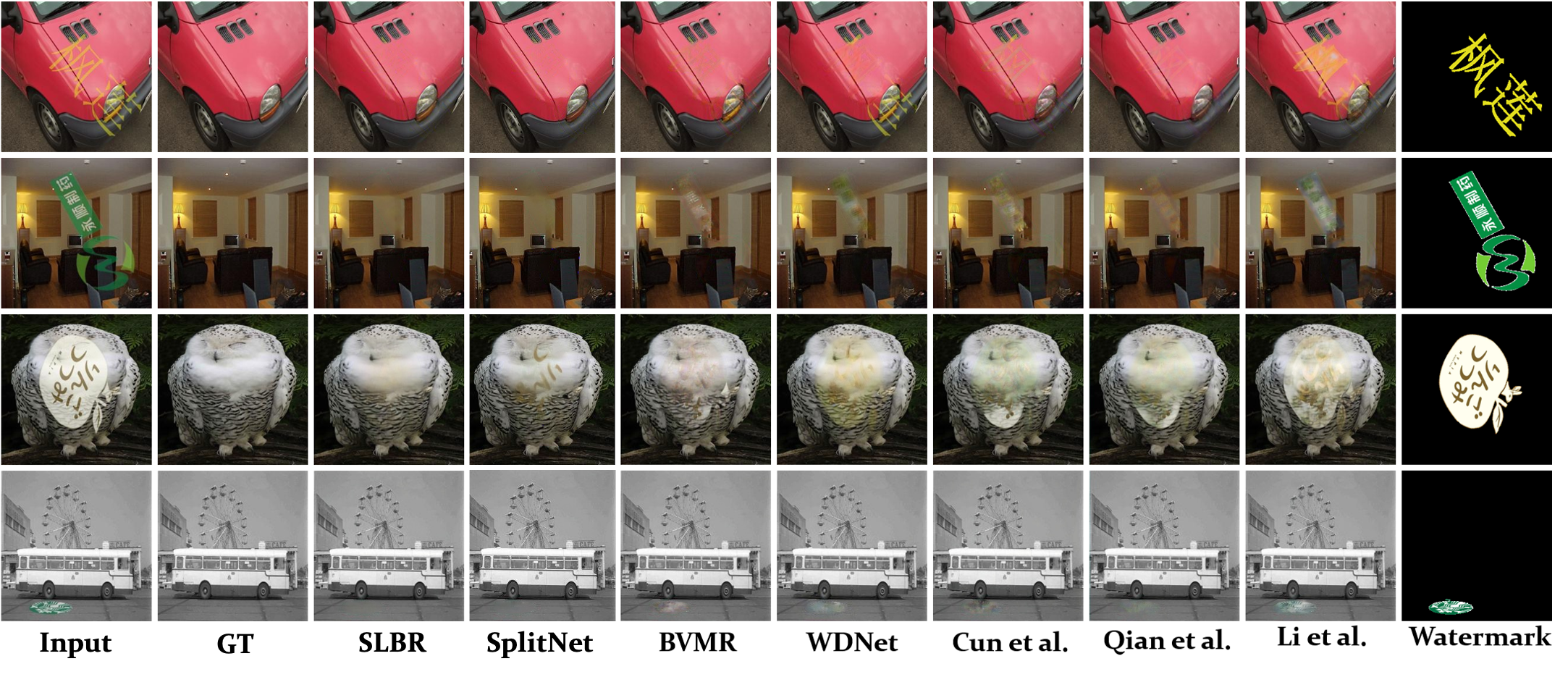} 
    \caption{Visualization results of different methods on CLWD~\cite{liu2021wdnet} dataset. Input is the watermarked image, GT is the ground-truth watermark-free image.}
    \label{fig:comparison}
\end{figure*}

\section{Experiments}
In this section, we first introduce our used datasets, implementation details, and evaluation metrics. Then, we compare our SLBR method with existing watermark removal methods and image content removal methods. We also provide visualization results of all methods to demonstrate the effectiveness of our method. Moreover, we conduct comprehensive ablation studies to investigate the benefit of each stage and each module in our network.

\subsection{Datasets and Implementation Details}
Following ~\cite{liu2021wdnet}, we conduct experiments on two large-scale benchmark datasets for watermark removal: Large-scale Visible Watermark
Dataset (LVW)~\cite{cheng2018large} and Colored Large-scale Watermark
Dataset (CLWD)~\cite{liu2021wdnet}.  LVW mainly
contains gray-scale watermarks, which have monotonous patterns and limited shapes. To overcome the shortcoming of LVW, the recent work~\cite{liu2021wdnet} contributed a large-scale dataset CLWD with colored and diverse watermarks, which is more realistic and challenging than LVW dataset.

\noindent\textbf{LVW~\cite{cheng2018large}:}
LVW contains 48,000 watermarked images made of 64 gray-scale watermarks for training and 12,000 watermarked images made of 16 gray-scale watermarks for testing.
The background images used in the training and test sets are randomly chosen from the train/val and test sets in PASCAL VOC2012 dataset~\cite{everingham2015pascal} respectively. 
 
\noindent\textbf{CLWD ~\cite{liu2021wdnet}:}
CLWD contains 60,000 watermarked images made of 160 colored watermarks for training and 10,000 watermarked images made of 40 colored watermarks for testing.  In CLWD, the watermarks are collected from open-sourced logo images websites. The original images used in training set and test sets are randomly chosen from PASCAL VOC2012~\cite{everingham2015pascal} training and test dataset respectively.  When making watermarked image, the transparency is set in the range of (0.3, 0.7). Besides, the size, locations, rotation angle, and transparency of each watermark is randomly set in different images. 

We implement our method using Pytorch~\cite{paszke2019pytorch}. We conduct all the experiments on the above two datasets. We set the input image size as $256 \times 256$. We choose Adam~\cite{kingma2014adam} optimizer with the initial learning rate  $0.001$, batch size $8$, and momentum parameters $\beta_1 = 0.5, \beta_2 = 0.999$. The hyper-parameters $\lambda_{\text{vgg}}$ and $\lambda_\text{mask}$ in (\ref{eqn:L_total}) are empirically set as $0.001$ and $1$ respectively, after a few trials by observing the quality of predicted masks and reconstructed images. 

\subsection{Baselines}
To the best of our knowledge, there are only a few deep learning methods specifically designed for watermark removal: conditional GAN based watermark removal method Li \emph{et al.}~\cite{li2019towards}, self-attention model Cao \emph{et al.}~\cite{cao2019generative}, blind visual motif removal method (BVMR)~\cite{hertz2019blind}, split and refine network(SplitNet)~\cite{cun2020split},watermark-decomposition network (WDNet)~\cite{liu2021wdnet}. We compare with these methods as the first group of baselines.
Following \cite{liu2021wdnet}, we also consider some image content removal methods and general image-to-image translation methods as the second group of baselines.
Concretely, we compare with attentive recurrent network Qian \emph{et al.}~\cite{qian2018attentive} for deraining, attention-guided dual hierarchical aggregation network Cun \emph{et al.}~\cite{cun2020towards} for shadow removal, and U-Net~\cite{ronneberger2015u} for general image-to-image translation. 

\subsection{Evaluation Metrics}
Following \cite{liu2021wdnet}, we adopt Peak Signal-to-Noise Radio (PSNR),  Structural Similarity (SSIM) \cite{wang2004image}, Root-Mean-Square (RMSE) distance, weighted Root-Mean-Square distance ($\text{RMSE}_w$) as evaluation metrics. The difference between RMSE and $\text{RMSE}_w$ lies in that $\text{RMSE}_w$ is only computed within the watermarked area. 

\begin{table*}[t] 
\centering
  \begin{tabular} {c|c|c|c|c| c c c c }
    \toprule[1pt]
 \textbf{\#} & \textbf{SMR} &\textbf{MBE} & \textbf{CFF} & \textbf{Skip-stage} & \multicolumn{4}{c}{\textbf{Evaluation Metrics}} \\ \cline{6-9}
    & & &  & &\multicolumn{1}{c|}{PSNR $\uparrow$}  &  \multicolumn{1}{c|}{SSIM $\uparrow$} & \multicolumn{1}{c|}{RMSE $\downarrow$} &  \multicolumn{1}{c}{RMSEw $\downarrow$}
    \\    \hline \hline
    1 & $\circ$ & $\circ$ & - & - &35.99 & 0.9708 &5.01 & 18.84 \\
    2 & $\times 1$ & $\circ$ & - & - & 36.38 & 0.9740 & 4.87 & 17.43 \\
    3 & $\times 3$ & $\circ$ & - & - & 36.50 & 0.9754 & 4.67 & 17.16 \\
    4 & $\times 3$ & $\times 1$ & - & - & 36.77 &  0.9759 & 4.53 & 16.92 \\
    5 & $\times 3$ & $\times 3$  & -& - & 36.90 & 0.9761 & 4.48  & 16.31 \\
    6 & $\times 3$ & $\times 3$  & $\times 0$ & - & 37.19  &0.9771 & 4.39 & 15.90  \\
    7 & $\times 3$ & $\times 3$  & $\times 1$ & - & 37.27  &0.9774 & 4.31 & 15.72  \\
    8 & $\times 3$ & $\times 3$  & $\times 2$ & - & 37.35  &0.9780 & 4.28 & 15.59 \\
    9 & $\times 3$ & $\times 3$  & $\times 3$ & - & 37.42 & 0.9785 & 4.24 &15.37\\
    10 & $\times 3$ & $\times 3$  & $\times 3$ & $\times 1$  &37.84 & 0.9797 & 3.94 &14.59\\
    11 & $\times 3$ & $\times 3$  & $\times 3$ & $\times 2$  &38.02 & 0.9801 & 3.84 &14.27\\
    12 & $\times 3$ & $\times 3$  & $\times 3$ & $\times 3$  &\textbf{38.28} & \textbf{0.9814} & \textbf{3.76} &\textbf{14.07}\\
     13 & $\times 3$ & $\times 3$  & $*$ & $\times 3$   & 37.65 & 0.9791 &4.07 & 14.87 \\
    \bottomrule[1pt]
  \end{tabular}
\caption{Ablation studies of our method on CLWD~\cite{liu2021wdnet}  dataset. $\circ$ means using original decoder block. $-$ means not using certain module or connection. $*$ means replacing CFF modules with original decoder blocks. $\times N$ means the times of using certain module or connection. For $\times 1$ (\emph{resp., $\times 2$}), we replace the module or add the skip-stage connection in the shallowest one (\emph{resp.}, two) layer(s). The best results are denoted in boldface.}
\label{tab:exp_ablation}
\end{table*}

\subsection{Experimental Results}
The results of all methods on two datasets are summarized in Table~\ref{tab:exp_baseline}. 
We reproduce the baseline results using their released code~\cite{qian2018attentive, cun2020towards, liu2021wdnet, hertz2019blind,ronneberger2015u} or our own implementation~\cite{li2019towards, cao2019generative}. One may notice that our reported results are different from those reported in \cite{liu2021wdnet}, especially the result of WDNet which is much worse than that in \cite{liu2021wdnet}. The performance degradation is attributed to a bug\footnote{They ignore the fact that return format of ``imread" function in OpenCV is unsigned, which will raise a numeric overflow issue when subtracting images.} in their released evaluation code. After fixing this bug, we re-evaluate and report the results of WDNet trained from scratch using their released code. 
One observation is that results on LVW dataset are much better than those on CLWD dataset, because LVW dataset only contains gray-scale watermarks and is much easier than CLWD dataset.
Another observation is that the image content removal methods~\cite{cun2020towards, qian2018attentive} and watermark removal methods~\cite{hertz2019blind,cun2020split, liu2021wdnet} based multi-task learning outperform image-to-image translation method~\cite{li2019towards} by a large margin, which verifies the  effectiveness and necessity of predicting watermark mask. Moreover, baselines SplitNet~\cite{cun2020split}, BVMR~\cite{hertz2019blind} and WDNet~\cite{liu2021wdnet} specifically designed for watermark removal perform more favorably on two datasets than image content removal methods~\cite{cun2020towards, qian2018attentive}.

Our SLBR method outperforms all baselines and achieves the best results on two datasets, which demonstrates the effectiveness of cross-task cross-stage information sharing and our devised modules.
Our performance gain on LVW dataset~\cite{cheng2018large} is not so obvious as that on CLWD dataset, which is again due to the simplicity of LVW dataset. In particular, gray-scale watermark removal task is much easier and we observe that the baseline methods can also capture the key pattern in LVW dataset within several training epochs. Therefore, the results on CLWD dataset can better justify the advantage of our proposed method.

For qualitative comparison, we show the visualization results of our method as well as baselines~\cite{cun2020split, hertz2019blind, liu2021wdnet, cun2020towards, qian2018attentive, li2019towards} in Figure~\ref{fig:comparison}. In each row, from left to right, we show the input watermarked image, the ground-truth watermark-free image, the watermark-free images generated by different methods, and the watermark. It can be seen that our method can reconstruct the structure information and texture details of background more clearly and coherently, which shows the advantage of our proposed method for watermark removal task. For example, in the first row, baseline methods are capable of removing the main part of watermark, but there are still some  remaining watermark, especially at the car light. In the second row,  baseline methods suffer from color inconsistency and noticeable artifacts. In contrast, our method can generally erase the entire watermark and reconstruct the background image with clear texture.


\begin{figure*}[t]
    \centering
    \includegraphics[width=0.9\textwidth]{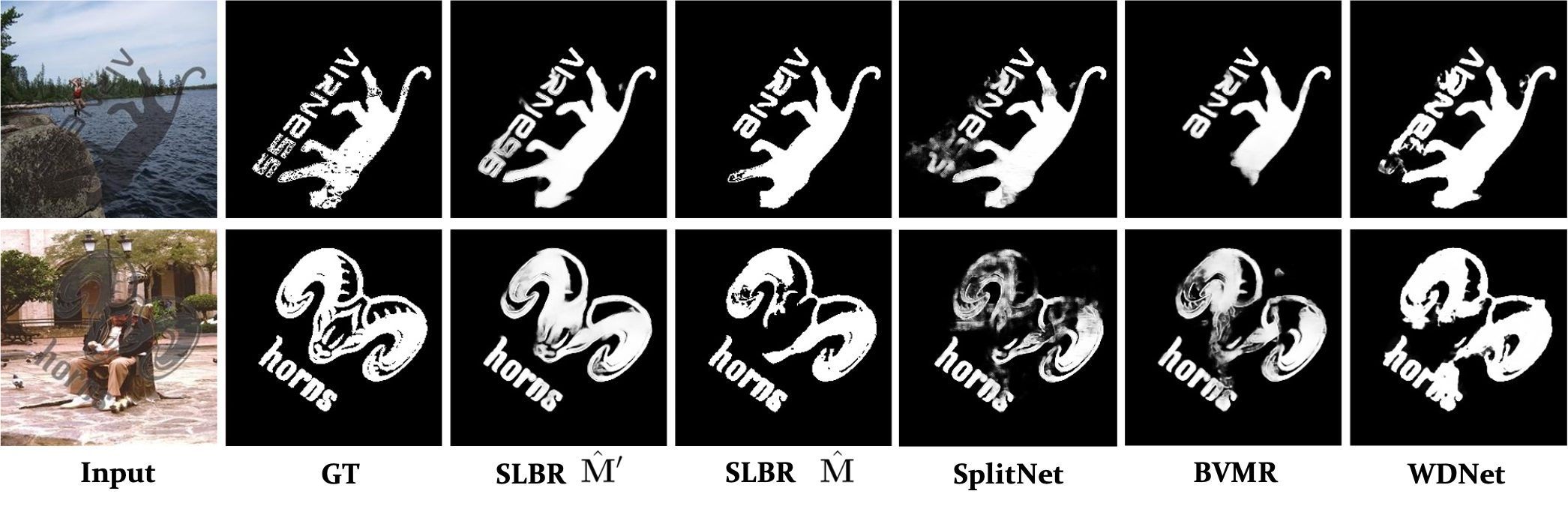} 
    \caption{Watermark localization results. From left to right, we show the input watermarked image, ground-truth watermark mask, the predicted results of our $\hat{\mathbf{M}}'$,  $\hat{\mathbf{M}}$, and baselines.}
    \label{fig:mask}
\end{figure*}

\begin{table}[t] 
\centering
  \begin{tabular} { c | c | c  }
    \toprule[1pt]
  \textbf{Method} & \multicolumn{2}{c}{\textbf{Evaluation Metrics}} \\ \cline{2-3}
    & \multicolumn{1}{c|}{F1} &  \multicolumn{1}{c}{IoU (\%)}
    \\    \hline \hline
    BVMR~\cite{hertz2019blind} & 0.7871 &  70.21 \\
    WDNet~\cite{liu2021wdnet} & 0.7240 &  61.20 \\
    SplitNet~\cite{cun2020split} & 0.8027 & 71.96 \\
    SLBR ($\hat{\mathbf{M}}$) & 0.8107  &  73.10 \\
    \textbf{SLBR} ($\hat{\mathbf{M}}'$) & \textbf{0.8234} &  \textbf{74.63} \\
    \bottomrule[1pt]
  \end{tabular}
\caption{Quantitative evaluation of watermark masks predicted by our method and baselines on CLWD~\cite{liu2021wdnet} dataset. }
\label{tab:mask}
\end{table}

\subsection{Ablation Studies}
In this section, we perform ablation studies to investigate the effectiveness of each module and each stage in our network. We start from a simple coarse stage network and gradually build up our full model. 
First, we only use the coarse stage and discard the refinement stage. Besides, we replace SMR and MBE modules with original decoder blocks~\cite{hertz2019blind} as mentioned in Section \ref{sec:fine_stage}. In this case, we obtain a standard U-Net structure except two separate decoder branches for watermark localization and watermark removal respectively. The results of this simplest case are reported in row 1 in Table \ref{tab:exp_ablation}. 

Then, we replace the last decoder block in the mask decoder branch with SMR, which corresponds to row 2 in Table \ref{tab:exp_ablation}. $\times 1$ means that we only use one SMR module. Furthermore, we replace all the decoder blocks in the mask decoder branch with SMR, corresponding to row 3 in Table \ref{tab:exp_ablation}. By comparing the first three rows in Table \ref{tab:exp_ablation}, it is evident that our SMR is better than original decoder block and able to predict the watermark mask more accurately. Besides, using three SMR works better than only using one SMR, which implies that learning better side output masks can contribute to better intermediate decoder features. 

Based on row 3, we replace the last decoder block in the background decoder branch with our MBE module, which import the output mask from mask decoder branch to enhance the last decoder feature, leading to the results in row 4. Furthermore, we replace all the decoder blocks in the background decoder branch with MBE, which utilizes all side output masks to enhance all the decoder features in the background decoder branch. The results using all three MBE modules are reported in row 5. By comparing row 3-5 in Table \ref{tab:exp_ablation}, we observe that our MBE is better than original decoder block and able to recover the background image better. Besides, using all three MBE performs better than only using one MBE, which implies that the side output masks from mask decoder branch can also benefit the reconstruction of watermark-free images.

Based on row 5, we introduce the refinement stage which only uses three encoder blocks and the final $1\times 1$  conv layer without CFF block, resulting in row 6 in Table \ref{tab:exp_ablation}. Then, we gradually increase the number of CFF blocks, leading to row 7-9 in Table \ref{tab:exp_ablation}. 
By comparing row 6-9,  we can draw a  conclusion that using CFF to aggregate multi-level features is necessary and using more CFF leads to better results. 

Based on row 9, we further bridge the coarse stage and the refinement stage by adding skip-stage connections. In row 10, we only connect the last decoder feature in the coarse stage and the first encoder feature in the refinement stage. In row 11-12, we link the last two (\emph{resp.} three) decoder features and the first two (\emph{resp.} three) encoder features using skip-stage connections, gradually yielding our full-fledged model. By comparing row 9-12,  we can observe that the information propagation through skip-stage connection is beneficial and more skip-stage connections can bring larger performance improvement. Finally, we replace CCF modules with decoder blocks~\cite{hertz2019blind}, making the refinement network a U-Net structure. The results are listed in row 13, based on which our design of refinement stage performs more favorably than a U-Net network structure.


\subsection{Watermark Localization}
In this section, we evaluate the quality of our predicted watermark masks $\hat{\mathbf{M}}$ and $\hat{\mathbf{M}}'$. 
We also compare with SplitNet~\cite{cun2020split}, BVMR~\cite{hertz2019blind}, WDNet~\cite{liu2021wdnet}, which can also predict watermark mask as a byproduct.
In terms of quantitative comparison, we calculate $F_1$ and IoU score based on the predicted mask and the ground-truth mask, where we simply use 0.5 as the threshold in all the experiments. The results are recorded in Table \ref{tab:mask}, which shows that $\hat{\mathbf{M}}'$ indeed improves $\hat{\mathbf{M}}$ and also outperforms the baselines~\cite{cun2020split, hertz2019blind,liu2021wdnet} by a large margin. 

We also show the predicted masks and ground-truth masks in Figure \ref{fig:mask} for qualitative comparison. From Figure \ref{fig:mask}, we can see that $\hat{\mathbf{M}}'$ is more complete and accurate. For example, in the first row, some texts in the rough estimation $\hat{\mathbf{M}}$ are missing. Thanks to our SMR block, the final result $\hat{\mathbf{M}}'$ is capable of predicting a complete mask, while other methods are struggling with the missed detection issue.


\section{Conclusion}
In this paper, we have studied watermark removal task and developed a two-stage multi-task network with novel MBE, SMR, and CCF modules, which can localize the watermark and recover the watermark-free image simultaneously. 
Extensive experiments on two datasets have verified the superiority of our proposed network.

\section*{Acknowledgement}
The work is supported by the National Key R\&D Program of China (2018AAA0100704) and is partially sponsored by National Natural Science Foundation of China (Grant No.61902247) and the Shanghai Science and Technology RD Program of China (20511100300). This work is also sponsored by Shanghai Municipal Science and Technology Major Project (2021SHZDZX0102).
\bibliographystyle{ACM-Reference-Format}
\bibliography{main}



\end{document}